\documentclass[runningheads]{llncs}
\usepackage[T1]{fontenc}
\usepackage{graphicx} % Required for inserting images
\usepackage{bm} 
\usepackage{amsmath}
\usepackage{amssymb}
\usepackage{mathtools}

\def \vector#1{\bm{#1}}
\def \matrix#1{\bm{#1}}

\usepackage{color}

\begin{document}

\title{Adjoint Sensitivities of Chaotic Flows without Adjoint Solvers: A Data-Driven Approach\thanks{This research has received financial support from the ERC Starting Grant No. PhyCo 949388 and UKRI AI for Net Zero grant EP/Y005619/1. L.M. is also grateful for the support from the grant EU-PNRR YoungResearcher TWIN ERC-PI\_0000005.}}
\titlerunning{Adjoint Sensitivities of Chaotic Flows without Adjoint Solvers}
\author{Defne E. Ozan \inst{1}\orcidID{0009-0006-4208-9981} \and 
Luca Magri \inst{1,2,3} \orcidID{0000-0002-0657-2611} }
\authorrunning{D.E. Ozan, L. Magri}
\institute{
Imperial College London, Department of Aeronautics, Exhibition Road, London SW7 2BX, UK \\
\email{d.ozan@imperial.ac.uk, l.magri@imperial.ac.uk}\\
\and
The Alan Turing Institute, London NW1 2DB, UK \and
Politecnico di Torino, DIMEAS, Corso Duca degli Abruzzi, 24 10129 Torino, Italy
}

\maketitle
\begin{abstract}
In one calculation, adjoint sensitivity analysis provides the gradient of a quantity of interest with respect to all system's parameters. Conventionally, adjoint solvers need to be implemented by differentiating computational models, which can be a cumbersome task and is code-specific. To propose an adjoint solver that is not code-specific, we develop a data-driven strategy.
We demonstrate its application on the computation of gradients of long-time averages of chaotic flows. First, we deploy a parameter-aware echo state network (ESN) to accurately forecast and simulate the dynamics of a dynamical system for a range of system's parameters. Second, we derive the adjoint of the parameter-aware ESN. Finally, we combine the parameter-aware ESN with its adjoint version to compute the  sensitivities to the system parameters. We showcase the method on a prototypical chaotic system. Because adjoint sensitivities in chaotic regimes diverge for long integration times, we analyse the application of ensemble adjoint method to the ESN. We find that the adjoint sensitivities obtained from the ESN match closely with the original system. This work opens possibilities for sensitivity analysis without code-specific adjoint solvers.
\end{abstract}

\section{Introduction}
Many computational applications including optimization, data assimilation, and uncertainty quantification require the information about how a quantity of interest is influenced by the system's parameters and initial conditions~e.g.,~\cite{magri2019AdjointMethodsDesign}. Precisely, sensitivity analysis provides the gradient of an objective functional with respect to all system's parameters. A direct method for estimating the sensitivity is to perturb the parameter and approximate the gradient with a finite difference between the base and the perturbed solutions. This procedure suffers from numerical errors and needs to be repeated for each parameter. Therefore, for systems with many parameters, the computational cost increases linearly with the number of parameters. Adjoint methods turn this around.  With an adjoint formulation, the sensitivity to all parameters is computed with a single operation, which requires the adjoint system. 

Whilst the adjoint is a powerful tool, there are some practical limitations to its use. The computation of the adjoint sensitivity relies on (i) the system equations to be known, and (ii) the system to be linearized, i.e., the computation of the Jacobian~\cite{giles2000IntroductionAdjointApproach,peter2010NumericalSensitivityAnalysis}. First, the adjoint sensitivity is only as accurate as the model assumptions. Second, the system might be high-dimensional and nonlinear, and in some cases the computational code might not be differentiable, all of which make the derivation of the Jacobian cumbersome. To address the challenges associated with developing adjoint solvers, we propose a data-driven approach, and apply this to a prototypical chaotic flow, which is a qualitative example of complex behaviour such as turbulence.

In chaotic systems, in which nearby trajectories diverge at an exponential rate, the adjoint system becomes unstable and the sensitivities diverge after some integration time. This poses a challenge as the quantities of interest are usually long-term time-averages. A possible approach that we also exploit in this paper is to take an ensemble of short-term trajectories~\cite{lea2000SensitivityAnalysisClimate}.

The objective of this paper is two-fold; (i) we introduce the parameter-aware echo state network to learn the parametrized dynamics of a chaotic system with, and (ii) we infer the sensitivities of an objective functional to the system parameters from the adjoint of parameter-aware echo state network.

\section{Parameter-aware Echo State Network}
Recurrent neural networks (RNNs) are specialized neural network architectures for sequential data processing and have been successfully employed for time series prediction. RNNs can be treated as discrete differential equations and can be studied with dynamical systems theory. In this work, we develop an RNN that learns the parametrized dynamics of a system such that we can make predictions in regimes that have not been seen during training. We further show that such a network can also infer the sensitivity of the dynamics to the parameters. We utilize the echo state network (ESN)~\cite{lukosevicius2012PracticalGuideApplying}, which whilst being a universal approximator~\cite{grigoryeva2018EchoStateNetworks} has the advantage over other RNNs of not needing backpropagation during training, and thus requiring less computational effort.  ESNs can make time-accurate short-term predictions~\cite{pathak2018ModelFreePredictionLarge,doan2020PhysicsinformedEchoState,racca2021RobustOptimizationValidation}, and infer long-term statistics and invariant properties, such as Lyapunov exponents, of chaotic flows~\cite{pathak2017UsingMachineLearning,margazoglou2023StabilityAnalysisChaotic}. A parameter-aware extension of the ESN has been formulated, and been applied to predict amplitude death, i.e., when the system bifurcates to fixed point solutions,~\cite{xiao2021PredictingAmplitudeDeath}, and multi-stable regimes including chaos~\cite{roy2022ModelfreePredictionMultistability}. The parameter-aware ESN is a nonlinear discrete map from reservoir state at time step $i$ to reservoir state at time step $i+1$
\begin{equation}
    \vector{r}(i+1) = (1-\alpha)\vector{r}(i) + \alpha\tanh(\matrix{W}_{in}[\vector{y}_{in}(i); \mathrm{diag}(\vector{\sigma}_p)(\vector{p}-\vector{k}_p)]+\matrix{W}\vector{r}(i)),
    \label{eq:parameter_esn_step}
\end{equation}
where $\vector{y}_{in}(i) \in \mathbb{R}^{N_y}$ is the input vector, $\vector{p} \in \mathbb{R}^{N_p}$ is the parameter vector, $\vector{r}(i) \in \mathbb{R}^{N_r}$ is the reservoir state, $\matrix{W}_{in} \in \mathbb{R}^{N_r \times (N_y + N_p)}$ is the input matrix, and $\matrix{W} \in \mathbb{R}^{N_r \times N_r}$ is the state matrix, and $\mathrm{diag}(\cdot)$ denotes a diagonal matrix that has $(\cdot)$ as its diagonal. The output is predicted from the reservoir state with a linear read-out layer
\begin{equation}
    \vector{\hat{y}}(i+1) = \matrix{W}_{out}\vector{r}(i+1),
    \label{eq:esn_readout}
\end{equation}   
$\vector{\hat{y}}(i+1) \in \mathbb{R}^{N_y}$ is the output vector, the dynamics of which we aim to model, and $\matrix{W}_{out} \in \mathbb{R}^{N_y \times N_r}$ is the output matrix. The matrices $\matrix{W}_{in}$ and $\matrix{W}$ are sparse, randomly generated, and not trained, whilst $\matrix{W}_{out}$ is trained via ridge regression~\cite{lukosevicius2012PracticalGuideApplying}. Training an ESN involves searching for optimal hyperparameters that consist of input matrix scaling $\sigma_{in}$, the spectral radius of the state matrix $\rho$, the leak rate $\alpha$, the Tikhonov regularizer $\lambda$, and finally two additional hyperparameters per parameter $p_i$; $k_p$ that shifts the parameter and $\sigma_p$ that scales it.

\section{Adjoint of Echo State Network}
We mathematically derive the adjoint of the ESN, which is an autonomous dynamical system when it runs in closed-loop. The dynamics~\eqref{eq:parameter_esn_step} can be expressed as a constraint $\vector{F}(i) = \vector{F}(\vector{r}(i),\vector{r}(i-1),\vector{p}) = 0$. We consider the time-averaged objective functional given as a sum over $N$ discrete time steps 
\begin{equation}
    \mathcal{J} = \frac{1}{N}\sum_{i = 1}^{N} \tilde{\mathcal{J}}(\vector{r}(i)),
    \label{eq:objective}
\end{equation}
in which the reservoir state $\vector{r}$ depends on the system's parameters $\vector{p}$. The goal of sensitivity analysis is to determine the gradient of the objective functional with respect to the system's parameters. For the parameter-aware ESN, the sensitivity is then expressed as
\begin{equation}
    \frac{d\mathcal{J}}{d\vector{p}} = \frac{1}{N}\sum_{i=1}^{N}\frac{d \tilde{\mathcal{J}}(\vector{r}(i))}{d \vector{r}(i)}\matrix{Q}(i),
    \label{eq:tangent_linear}
\end{equation}
where we define $\matrix{Q}(i) \coloneqq d\vector{r}(i) / d\vector{p}, \; \matrix{Q}(i) \in \mathbb{R}^{N_r \times N_p}$. Notice that the dimension of $\matrix{Q}(i)$ grows with the number of parameters. This increasing computational cost motivates solving the adjoint problem instead. The Lagrangian of the objective functional, $\mathcal{J}$, subjected to the system dynamics, $\vector{F} = 0$, is
\begin{equation}
    \mathcal{L} \coloneqq \mathcal{J} - \langle\vector{q}^+, \vector{F}\rangle,
    \label{eq:lagrangian}
\end{equation}
where the dot product $\langle\vector{q}^+, \vector{F}\rangle$ is defined as
\begin{equation}
    \langle\vector{q}^+, \vector{F}\rangle \coloneqq \sum_{i = 1}^N \vector{q}^{+T}(i) \vector{F}(i),
\end{equation}
and $\vector{q}^+ \in \mathbb{R}^{N_r}$ are the Lagrange multipliers, or the adjoint variables. 
We solve for the gradient $d\mathcal{L}/d\vector{p} = d\mathcal{J}/d\vector{p}$.
% \begin{equation}
%     \begin{split}
%     \frac{d\mathcal{L}}{d\vector{p}} = &\frac{1}{N}\sum_{i = 1}^N \frac{\partial\tilde{\mathcal{J}}}{\partial\vector{r}(i)}\frac{d\vector{r}(i)}{d\vector{p}} \\
%     &- \sum_{i = 1}^N 
%     \vector{q}^{+T}(i)
%     \left(
%         \frac{\partial \vector{F}(i)}{\partial \vector{p}}
%         + \frac{\partial \vector{F}(i)}{\partial \vector{r}(i)}\frac{d\vector{r}(i)}{d\vector{p}}
%         + \frac{\partial \vector{F}(i)}{\partial \vector{r}(i-1)}\frac{d\vector{r}(i-1)}{d\vector{p}}
%     \right). 
%      \end{split}
% \end{equation}
After expanding the summation and rearranging the terms, we choose the Lagrange multipliers such that we can eliminate the terms $d\vector{r}(i)/d\vector{p}$,~e.g.,~\cite{magri2019AdjointMethodsDesign}. This provides us with the evolution equations of the adjoint variables
\begin{subequations}\label{eq:adjoint}
    \begin{align}
        \frac{d\mathcal{J}}{d\vector{p}} &= \sum_{i = 1}^N \vector{q}^{+T}(i) \frac{\partial \vector{r}(i)}{\partial \vector{p}}, \\
        \vector{q}^+(i) &=  \frac{1}{N}\frac{\partial \tilde{\mathcal{J}}(\vector{r}(i))}{\partial \vector{r}(i)}^T + \frac{\partial\vector{r}(i+1)}{\partial \vector{r}(i)}^T\vector{q}^+(i+1), \\
        \vector{q}^+(N) &= \frac{1}{N}\frac{\partial \tilde{\mathcal{J}}(\vector{r}(N))}{\partial \vector{r}(N)}^T. \label{eq:adjoint_terminal}
    \end{align}
\end{subequations}
Practically, we first let the ESN run autonomously for the given time window and save the direct solution, which serves as the base trajectory. We then solve the adjoint equations backwards in time starting from the terminal condition $\vector{q}^+(N)$~\eqref{eq:adjoint_terminal}. The solution of the adjoint equations requires the computation of the Jacobian evaluated at the base trajectory, i.e., the gradient of the reservoir state at time step $i+1$ with respect to the reservoir state at time step $i$ 
\begin{equation}
    \frac{\partial \vector{r}(i+1)}{\partial \vector{r}(i)} = (1-\alpha)\matrix{I}_{N_r \times N_r} + \alpha\mathrm{diag}(1-\tilde{\vector{r}}^2(i))(\matrix{W}_{in}^{y}\matrix{W}_{out}+\matrix{W}),
\end{equation}
and the gradient with respect to the parameters
\begin{equation}
    \frac{\partial \vector{r}(i+1)}{\partial \vector{p}} = \alpha \mathrm{diag}(1-\tilde{\vector{r}}^2(i)) \matrix{W}_{in}^{p}\mathrm{diag}(\vector{\sigma}_p), 
\end{equation}
where $\tilde{\vector{r}}(i) = (\vector{r}(i+1)-(1-\alpha)\vector{r}(i))/\alpha$, $\matrix{W}_{in} = [\matrix{W}_{in}^y \; \matrix{W}_{in}^p]$, and $\matrix{I}$ denotes the identity matrix.

\section{Computation of data-driven chaotic sensitivities}
We demonstrate the data-driven computation of adjoint sensitivities on the Lorenz 63 system, which is a reduced-order model to study atmospheric convection 
\begin{equation}
    \frac{dx}{dt} = s(y-x), \quad 
    \frac{dy}{dt} = x(r-z) - y, \quad
    \frac{dz}{dt} = xy-bz,
\label{eq:lorenz63}
\end{equation}
where $x, y, z$ are the state variables, and $s, r, b$ are the system's parameters.  
We generate the dataset for the training and validation of the ESN by time-marching the ODEs~\eqref{eq:lorenz63} for different sets of parameters. The parameters $(s = 10, r = 28, b = 8/3 \approx 2.667)$ lead to a chaotic solution~\cite{lorenz1963DeterministicNonperiodicFlow}.  We randomly choose 20 regimes for training, and 5 regimes for validation from a grid of parameters, $s = \{8,10,12,14,16\}$, $r = \{30, 35, 40, 45, 50\}$, and $b = \{1,1.5,2,2.5,3\}$.  We observe that it is important to validate the ESN on regimes unseen during the training in order to choose a model that generalizes well over a range of different parameters. The regimes in the training and validation datasets display chaotic behaviour with varying Lyapunov times (LTs), i.e., the time-scale of divergence of two nearby trajectories in a chaotic system,  (between 0.77 and 4.80 time units). The numerical integration is performed with a fourth-order Runge-Kutta scheme with a time-step of $\Delta t = 0.01$. After a transient, we select the first 4 time units for washout and 10 time units as training data. The hyperparameters for the ESN are determined via Bayesian optimization~\cite{racca2021RobustOptimizationValidation}, in which we evaluate the short-term closed-loop performance of random realisations of the model on the validation dataset. The optimal hyperparameters for an ESN of reservoir size, $N_r = 1200$ and a connectivity, $ N_{conn}= 3$ between the reservoir state variables are found as $\rho=0.2201$, $\sigma_{in}=0.0679$, $\sigma_{s}=0.0028$, $k_{s}=68.73$, $\sigma_{r}=0.0015$, $k_{r}=84.81$, $\sigma_{b}=0.0393$, $k_{b}=74.46$, $\alpha=0.8853$, and $\lambda=10^{-10}$.
% provided in Table~\ref{tab:hyperparameters}.

% \addtolength{\tabcolsep}{2pt} 
% \begin{table}\label{tab:hyperparameters}
%     \caption{Optimal hyperparameters of the Echo State Network with a reservoir size of $N_r = 1200$ and connectivity $N_{conn} = 3$ that infers the parametrized dynamics of Lorenz 63. \textcolor{red}{Can we put these parameters inline to save up precious space?}}
%     \begin{tabular}{cccccccccc}
%         $\rho$ & $\sigma_{in}$ & $\sigma_{s}$ & $k_{s}$ & $\sigma_{r}$ & $k_{r}$ & $\sigma_{b}$ & $k_{b}$ & $\alpha$ & $\lambda$ \\ \hline
%         % 0.01 -- 1 & 0.001 -- 1 & 0.001 -- 0.01 & -12 -- 12 & 0.001 -- 0.01 & -40 -- 40 & 0.01 -- 0.1 & -2 -- 2 & 0.1 -- 1 & $10^{-10}$ -- $10^{-3}$ \\ \hline
%         0.2201 & 0.0.0679 & 0.0028 & 68.73 & 0.0015 & 84.81 & 0.0393 & 74.46 & 0.8853 & $10^{-10}$
%     \end{tabular}

% \end{table}

\begin{figure}[h!]
    \includegraphics[width=\linewidth]{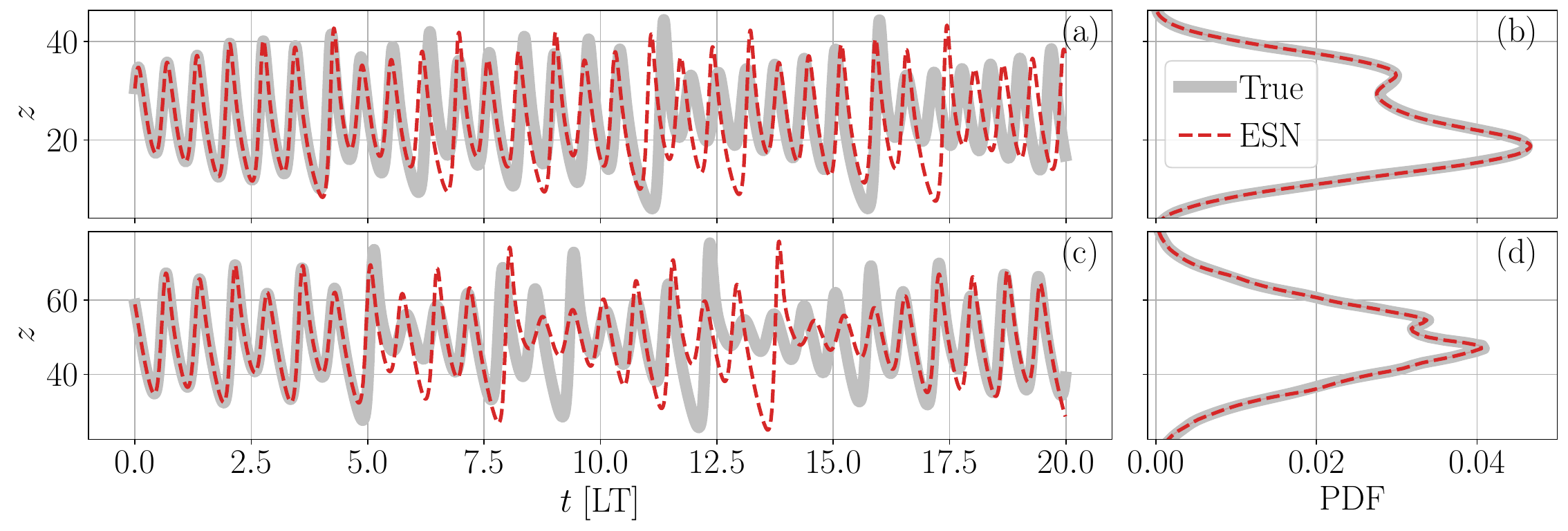}
    \caption{Short-term prediction (a,c) and long-term inference of the statistics (b,d) of $z$ for two regimes with different Lyapunov times (LTs); (a,b) $(s = 10, r = 28, b = 8/3 \approx 2.667)$, and (c,d) $(s = 13, r = 52, b = 1.75)$. The statistics are calculated over 5000 LTs, after a washout stage where we repeatedly feed the same initial condition, and a transient time that is discarded. The parameter-aware echo state network can successfully infer the dynamics and long-term statistics of different chaotic regimes even when they were not seen during training.}
    \label{fig:pred_stats}
\end{figure}

Previous works have applied and analysed the ensemble adjoint method to compute the sensitivity of the time-averaged $z$, $\bar{z}$, to the parameter $\rho$ in the Lorenz 63 system due to their nearly linear relationship~e.g.,~\cite{lea2000SensitivityAnalysisClimate,eyink2004RuelleLinearResponse}. Therefore, we  choose $\bar{z}$ as the quantity of interest. First, the short-term prediction and the long-term inference of statistics for $z$ are illustrated in Fig.~\ref{fig:pred_stats} for two regimes; (i) the reference configuration, $(s = 10, r = 28, b = 8/3 \approx 2.667)$ with a Lyapunov time of $LT = 1.1$, and (ii) a configuration with a shorter Lyapunov time $LT = 0.8$ $(s = 13, r = 52, b = 2.0)$. Neither of these regimes were seen during training or validation. Nonetheless, time-accurate short-term forecasting can be performed with a predictability horizon e.g.,~\cite{racca2021RobustOptimizationValidation} of $4.8\; LT$ for the regime (i), and $5.8\; LT$ for regime (ii) (average over 100 initial conditions on the attractor), and the long-term statistics are captured for a wide range of seen and unseen regimes with a single instance of the parameter-aware ESN.

% Mean PH: 4.7974825976013395 Std PH: 1.2754412938812416
% Mean PH: 5.81918979327503 Std PH: 1.2109050978672167

\begin{figure}[h!]
    \includegraphics[width=\linewidth]{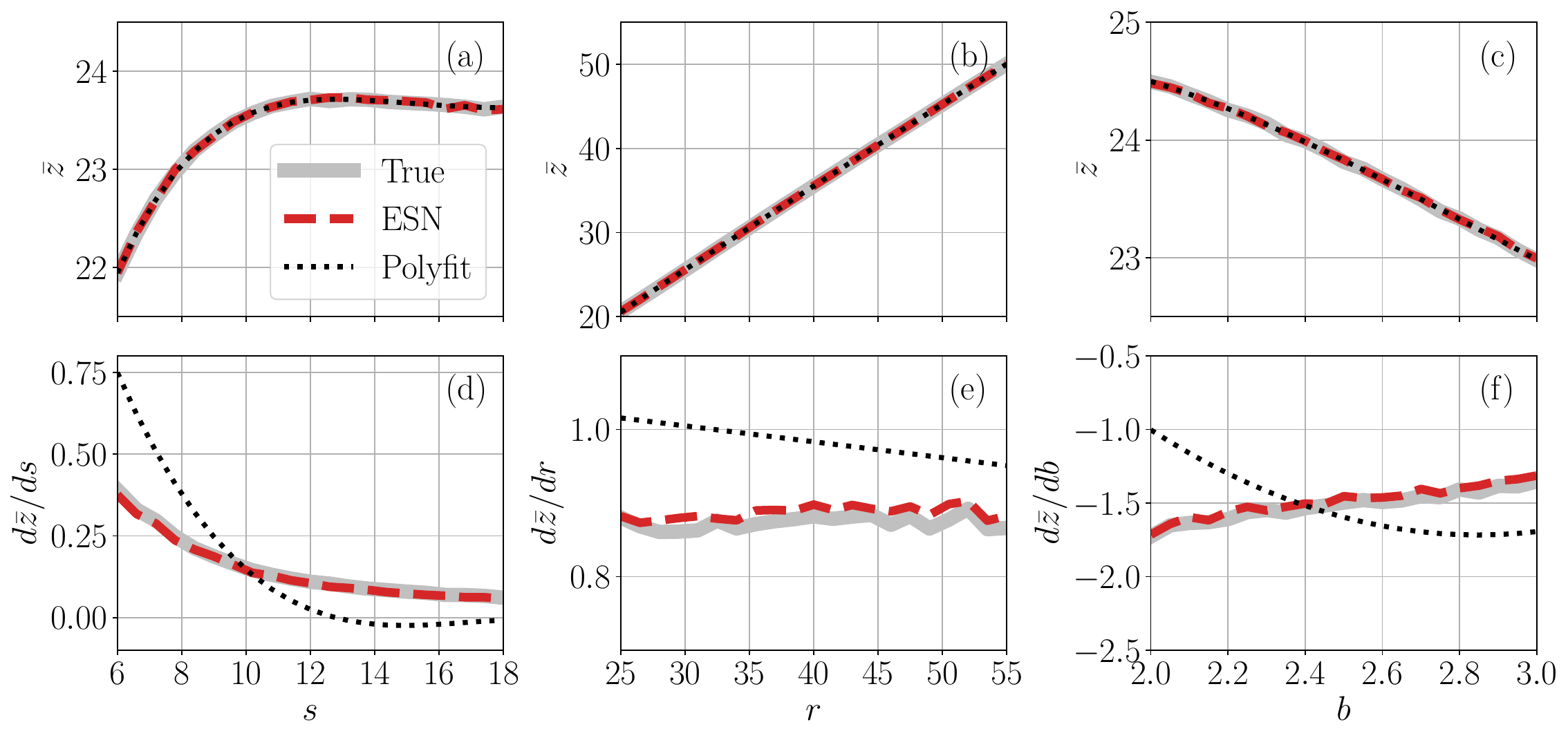}
    \caption{Inference of sensitivities of $\bar{z}$ to parameters, $s$, $r$, and $b$. Top row (a,b,c) shows the change of $\bar{z}$ for varying $s$, $r$, and $b$, for the true system, echo state network (ESN), and a polynomial fit on the true system's values. Botton row (d,e,f) shows the derivative of $\bar{z}$ with respect to the respective parameter obtained by the ensemble adjoint method using the true system's adjoint, ESN's adjoint, and by differentiating the polynomial fit.}
    \label{fig:sensitivity_1}
\end{figure}

Next, we compute the ``climate'' sensitivity of $\bar{z}$ to all parameters using an ensemble adjoint method~\cite{lea2000SensitivityAnalysisClimate}. Climate sensitivity refers to the long-term behaviour of the system, where the sensitivity is independent of initial condition. In the ensemble adjoint method, the sensitivities over an ensemble of short time-series with different initial conditions on the attractor are computed and the climate sensitivity is estimated by the mean of the ensemble. The sensitivity estimator is associated with a bias and a variance depending on the integration time of each trajectory~\cite{chandramoorthy2017AnalysisEnsembleAdjoint}. While the bias decreases with increasing integration time, the variance increases and the so-called L\'{e}vy flights, i.e., long jumps in the mean estimation, appear, requiring more ensemble members for convergence~\cite{eyink2004RuelleLinearResponse}. We opt for an integration time of half a Lyapunov time and an ensemble of 10000 trajectories, which provide an acceptable trade-off between accuracy and computational cost. Figure~\ref{fig:sensitivity_1} compares the objective and the adjoint sensitivity estimates of the ESN with the original system as well as a direct estimate using a polynomial fit of the objective values. Sensitivities pertaining to the parameters $s$, $r$, and $b$, are shown while each of them is varied and the other two are fixed at the given values for the regime (i). The ESN estimates closely match with the true estimates obtained via ensemble adjoint method. The difference between the direct estimate and the adjoint estimates is expected due to the above mentioned bias. We only show the component of the sensitivity associated with the parameter we vary, but in fact the adjoint method has output the sensitivities to all parameters, which similarly match with the true estimates. 

\section{Conclusions and future directions}
We propose a data-driven method to obtain adjoint sensitivities of a system with respect to its parameters using a parameter-aware extension of echo state network. The fact that an echo state network (ESN) is a dynamical system, for which an adjoint system exists, and the success of ESNs to not only forecast predictions and but also infer invariant properties, motivate this approach. We demonstrate this approach on a prototypical chaotic system with three system's parameters. First, we train a parameter-aware ESN on data from different chaotic regimes of Lorenz system, and show its performance when running autonomously to make short-term time-accurate predictions and replicate long-term statistics of the chaotic attractor at regimes that have not been seen during training or validation. Second, using the derived adjoint equations of the ESN, we compute the sensitivity of a time-averaged state variable, $\bar{z}$, at varying parameter values to all parameters, $s$, $r$, and $b$. We find that the estimated sensitivities match closely with the sensitivities obtained by the adjoint of the original system. We observe a known bias~\cite{chandramoorthy2017AnalysisEnsembleAdjoint} between the ensemble adjoint sensitivities and direct estimates calculated by fitting a polynomial to the objective function. This work opens possibilities for data-driven sensitivity analysis of chaotic flows without adjoint solvers. Future directions include analysis of the scalability of the method to higher-dimensional systems. 
%the adjoint sensitivity estimate, for which methods such as shadowing has been developed, and application of this method in conjunction with a data-driven reduced order model.

\bibliographystyle{splncs04}

\begin{thebibliography}{10}
    \providecommand{\url}[1]{\texttt{#1}}
    \providecommand{\urlprefix}{URL }
    \providecommand{\doi}[1]{https://doi.org/#1}
    
    \bibitem{chandramoorthy2017AnalysisEnsembleAdjoint}
    Chandramoorthy, N., Fernandez, P., Talnikar, C., Wang, Q.: An {{Analysis}} of
      the {{Ensemble Adjoint Approach}} to {{Sensitivity Analysis}} in {{Chaotic
      Systems}}. In: 23rd {{AIAA Computational Fluid Dynamics Conference}}.
      {American Institute of Aeronautics and Astronautics}, Denver, Colorado (Jun
      2017). \doi{10.2514/6.2017-3799}
    
    \bibitem{doan2020PhysicsinformedEchoState}
    Doan, N., Polifke, W., Magri, L.: Physics-informed echo state networks. Journal
      of Computational Science  \textbf{47},  101237 (Nov 2020).
      \doi{10.1016/j.jocs.2020.101237}
    
    \bibitem{eyink2004RuelleLinearResponse}
    Eyink, G.L., Haine, T.W.N., Lea, D.J.: Ruelle's linear response formula,
      ensemble adjoint schemes and {{L{\'e}vy}} flights. Nonlinearity
      \textbf{17}(5),  1867--1889 (Sep 2004). \doi{10.1088/0951-7715/17/5/016}
    
    \bibitem{giles2000IntroductionAdjointApproach}
    Giles, M.B., Pierce, N.A.: An {{Introduction}} to the {{Adjoint Approach}} to
      {{Design}}. Flow, Turbulence and Combustion  \textbf{65}(3/4),  393--415
      (2000). \doi{10.1023/A:1011430410075}
    
    \bibitem{grigoryeva2018EchoStateNetworks}
    Grigoryeva, L., Ortega, J.P.: Echo state networks are universal. Neural
      Networks  \textbf{108},  495--508 (Dec 2018).
      \doi{10.1016/j.neunet.2018.08.025}
    
    \bibitem{lea2000SensitivityAnalysisClimate}
    Lea, D.J., Allen, M.R., Haine, T.W.N.: Sensitivity analysis of the climate of a
      chaotic system. Tellus A  \textbf{52}(5),  523--532 (Oct 2000).
      \doi{10.1034/j.1600-0870.2000.01137.x}
    
    \bibitem{lorenz1963DeterministicNonperiodicFlow}
    Lorenz, E.N.: Deterministic {{Nonperiodic Flow}}. Journal of the Atmospheric
      Sciences  \textbf{20}(2),  130--141 (Mar 1963).
      \doi{10.1175/1520-0469(1963)020<0130:DNF>2.0.CO;2}
    
    \bibitem{lukosevicius2012PracticalGuideApplying}
    Luko{\v s}evi{\v c}ius, M.: A {{Practical Guide}} to {{Applying Echo State
      Networks}}. In: Montavon, G., Orr, G.B., M{\"u}ller, K.R. (eds.) Neural
      {{Networks}}: {{Tricks}} of the {{Trade}}, vol.~7700, pp. 659--686. Springer
      Berlin Heidelberg, Berlin, Heidelberg (2012).
      \doi{10.1007/978-3-642-35289-8_36}
    
    \bibitem{magri2019AdjointMethodsDesign}
    Magri, L.: Adjoint {{Methods}} as {{Design Tools}} in {{Thermoacoustics}}.
      Applied Mechanics Reviews  \textbf{71}(2),  020801 (Mar 2019).
      \doi{10.1115/1.4042821}
    
    \bibitem{margazoglou2023StabilityAnalysisChaotic}
    Margazoglou, G., Magri, L.: Stability analysis of chaotic systems from data.
      Nonlinear Dynamics  \textbf{111}(9),  8799--8819 (May 2023).
      \doi{10.1007/s11071-023-08285-1}
    
    \bibitem{pathak2018ModelFreePredictionLarge}
    Pathak, J., Hunt, B., Girvan, M., Lu, Z., Ott, E.: Model-{{Free Prediction}} of
      {{Large Spatiotemporally Chaotic Systems}} from {{Data}}: {{A Reservoir
      Computing Approach}}. Physical Review Letters  \textbf{120}(2),  024102 (Jan
      2018). \doi{10.1103/PhysRevLett.120.024102}
    
    \bibitem{pathak2017UsingMachineLearning}
    Pathak, J., Lu, Z., Hunt, B.R., Girvan, M., Ott, E.: Using machine learning to
      replicate chaotic attractors and calculate {{Lyapunov}} exponents from data.
      Chaos: An Interdisciplinary Journal of Nonlinear Science  \textbf{27}(12),
      121102 (Dec 2017). \doi{10.1063/1.5010300}
    
    \bibitem{peter2010NumericalSensitivityAnalysis}
    Peter, J.E., Dwight, R.P.: Numerical sensitivity analysis for aerodynamic
      optimization: {{A}} survey of approaches. Computers \& Fluids
      \textbf{39}(3),  373--391 (Mar 2010). \doi{10.1016/j.compfluid.2009.09.013}
    
    \bibitem{racca2021RobustOptimizationValidation}
    Racca, A., Magri, L.: Robust {{Optimization}} and {{Validation}} of {{Echo
      State Networks}} for learning chaotic dynamics. Neural Networks
      \textbf{142},  252--268 (Oct 2021). \doi{10.1016/j.neunet.2021.05.004}
    
    \bibitem{roy2022ModelfreePredictionMultistability}
    Roy, M., Mandal, S., Hens, C., Prasad, A., Kuznetsov, N.V., Dev~Shrimali, M.:
      Model-free prediction of multistability using echo state network. Chaos: An
      Interdisciplinary Journal of Nonlinear Science  \textbf{32}(10),  101104 (Oct
      2022). \doi{10.1063/5.0119963}
    
    \bibitem{xiao2021PredictingAmplitudeDeath}
    Xiao, R., Kong, L.W., Sun, Z.K., Lai, Y.C.: Predicting amplitude death with
      machine learning. Physical Review E  \textbf{104}(1),  014205 (Jul 2021).
      \doi{10.1103/PhysRevE.104.014205}
    
    \end{thebibliography}

\end{document}